\documentclass{ifacconf}

\usepackage{graphicx}      
\usepackage{natbib}        
\usepackage{amsmath,amssymb}
\usepackage[dvipsnames]{xcolor}
\usepackage{multirow}
\usepackage{eurosym}
\usepackage{tikz}

\newcommand{\Exp}{\mathbb{E}}
\newcommand{\cP}{\mathcal{P}}

\newcommand{\vehsum}{\sum_{i = 1}^N}
\newcommand{\timesum}{\sum_{k = 0}^{T-1}}

\newcommand{\timelist}{k = 0,\dots,T-1}
\newcommand{\dt}{\tau}

\newcommand{\dam}{\mathrm{dam}}
\newcommand{\asm}{\mathrm{asm}}
\newcommand{\veh}{\mathrm{veh}}

\newcommand{\posp}[1]{\left[#1\right]^+}
\newcommand{\negp}[1]{\left[#1\right]^-}

\newcommand{\edam}{e^\dam}
\newcommand{\ep}{e^+}
\newcommand{\en}{e^-}
\newcommand{\Dr}{\Delta_e^\circ}
\newcommand{\Dmax}{\Delta_e^{\max}}
\newcommand{\et}{\tilde{e}^\dam}

\graphicspath{{figures/}}

\newcommand\copyrighttext{%
	\small \begin{center} \color{red} \textcopyright\,2023 Elsevier. Personal use of this material is permitted. Permission from Elsevier must be obtained for all other uses, in any current or future media, including reprinting/republishing this material for advertising or promotional purposes, creating new collective works, for resale or redistribution to servers or lists, or reuse of any copyrighted component of this work in other works. \end{center}}
\newcommand\copyrightnotice{%
	\begin{tikzpicture}[remember picture,overlay]
		\node[anchor=south,yshift=27.6cm] at (current page.south) 
		{\color{red}\fbox{\parbox{\dimexpr\textwidth-\fboxsep-\fboxrule\relax}{\copyrighttext}}};
	\end{tikzpicture}%
}

\begin{document}

\begin{frontmatter}	
	\copyrightnotice
	\title{Vehicle-to-Grid and ancillary services:\\ a profitability analysis under uncertainty\thanksref{footnoteinfo}}
	
	\thanks[footnoteinfo]{This work has been funded by the European Union’s Horizon research and innovation programme FLOW under grant agreement no. 101056730 and by the Research Fund for the Italian Electrical System under the contract agreement between RSE S.p.A. and the Ministry of Economic Development - General Directorate for the Electricity Market, Renewable Energy and Energy Efficiency, Nuclear Energy in compliance with the Decree of April 16th, 2018.}
	
	\author[RSE]{Federico~Bianchi}
	\author[PoliMi]{Alessandro~Falsone}
	\author[BeP]{Riccardo~Vignali}
	
	\address[RSE]{Ricerca sul Sistema Energetico -- RSE S.P.A.,\\ Via R. Rubattino, 54, 20134, Milano, Italy\\ (e-mail: federico.bianchi@rse-web.it)}
	\address[PoliMi]{Dipartimento di Elettronica, Informazione e Bioingegneria,\\ Politecnico di Milano, Via Ponzio, 34/5, 20133, Milano, Italy.\\ (e-mail: alessandro.falsone@polimi.it)}
	\address[BeP]{Eni Plenitude S.p.A.,\\ Via Ripamonti 85, 20141, Milano, Italy.\\ (e-mail: riccardo.vignali@eniplenitude.com)}
	
	\begin{abstract}
		The rapid and massive diffusion of electric vehicles poses new challenges to the electric system, which must be able to supply these new loads, but at the same time opens up new opportunities thanks to the possible provision of ancillary services. Indeed, in the so-called Vehicle-to-Grid (V2G) set-up, the charging power can be modulated throughout the day so that a fleet of vehicles can absorb an excess of power from the grid or provide extra power during a shortage.
		To this end, many works in the literature focus on the optimization of each vehicle daily charging profiles to offer the requested ancillary services while guaranteeing a charged battery for each vehicle at the end of the day.
		However, the size of the economic benefits related to the provision of ancillary services varies significantly with the modeling approaches, different assumptions, and considered scenarios. In this paper we propose a profitability analysis with reference to a recently proposed framework for V2G optimal operation in presence of uncertainty. We provide necessary and sufficient conditions for profitability in a simplified case and we show via simulation that they also hold for the general case.
	\end{abstract}
	
	
\end{frontmatter}

\section{Introduction} \label{sec:introduction}
Electric Vehicles (EVs) sales continue to break records as nearly $10\%$ of global car sales were electric in 2021, four times the market share in 2019,~\cite{IEA2022}. This rapid and significant spread of EVs plays a fundamental role in the energy transition. In addition to the new challenges associated with charging needs, the introduction of EVs represents a new opportunity, thanks to the possible provision of ancillary services. Indeed, the modulation of the charging power, and even the discharge of the vehicles, enable the provision of services to the electricity network,~\cite{liu2013opportunities}. However, since an individual EV’s energy capacity is limited, EVs need to be grouped by means of EV aggregators in order to form a flexible load with enough energy content for grid operations. Once the fleet is formed, the aggregator has to coordinate the actions of the EV pool to participate in electricity markets, guaranteeing compliance with traded consumption plans and services,~\cite{bessa2012economic}.

The optimal dispatch of the charging power of each vehicle is usually formulated as an optimization problem, in which the aggregator aims to maximize its revenues from the provision of services, minimizing at the same time the EVs charging costs while satisfying the requests of the EV owners, \emph{e.g.,} minimum charging level at departure. Solving this problem needs to account for several factors: aggregator business model, technical limitations of vehicles and aggregator, availability of vehicles, market outcomes. Realistic formulations of this problem necessarily involves taking into account uncertainty in the fleet behavior and energy markets. Specifically, it is necessary to model the random presence of vehicles, the initial uncertain State of Charge (SOC) with which the vehicles start parking and the actual service signal provided by the Transmission System Operator (TSO). Several techniques have been proposed in the last two decades for solving this optimization problem, which differ on how uncertainty is handled and the modeling choices for the discussed factors, see~\cite{garcia2014plug, tan2016integration, nimalsiri2019survey}, for a comprehensive review on different strategies.

In many of these works, costs-benefits analysis related to the provision of ancillary services have been made experimentally by, \emph{e.g.}, varying energy market prices and also battery degradation costs, from the point of view of both EV owners and aggregators,~\cite{sortomme2011optimal,de2012economic,calvillo2016vehicle}.
However, the size of the benefits varies significantly due to varying modeling approaches, different assumptions, considered applications, countries and vehicle types, often leading to inconsistent and contradictory results, ~\cite{heilmann2021factors}. Also, to the best of our knowledge, no-one provided an overall picture describing profitability conditions for offering balancing services. For example, one may be interested in assessing which should be the expected profit of an upward service in relation to the per unit energy cost used for charging the vehicle, in order to make this kind of service profitable. Clearly, answering to this question requires to account for the (uncertain) TSO signal, both in the modeling framework and in the costs-benefits analysis. This kind of analysis helps in incentivizing the participation of EVs' owners in the mentioned aggregation scheme, thus contributing to the transition that the energy sector is facing which asks for additional sources of flexibility to guarantee reliable grid operation.

Recently, a first attempt towards a comprehensive framework for optimal dispatching in presence of uncertainty has been made in~\cite{Vignali2022}. The authors considered all the three sources of uncertainty discussed above, adopting a robust paradigm to enforce the constraints and an expectation paradigm for the cost function. However, they did not analyze the profitability of providing ancillary services. In this work we build upon the framework introduced by~\cite{Vignali2022}, and we provide necessary and sufficient conditions for the profitability of offering upward and downward balancing services. Specifically, we provide analytic conditions in a simplified case and we show via simulation that they also hold for the general case. Based on these conditions, we also provide insights on how to make ancillary services more profitable by considering the possibility of participating also to intraday energy markets besides the day-ahead and ancillary service markets, so as to compensate for the effects of the actual provision of services during the day.


\section{Preliminaries} \label{sec:preliminaries}
Let us first briefly recall the framework introduced in~\cite{Vignali2022} and introduce two common scenarios we will consider in the profitability analysis.

\subsection{Framework} \label{sec:framework}
With reference to an aggregator of electric vehicles (EVs),~\cite{Vignali2022} deals with the finite horizon optimal control problem of planning the next day power exchange profile and maximum amount of upward and downward power variations the fleet is able to provide to the main grid, so as to minimize the EVs charging costs and maximize the revenues associated with the ancillary service provision.

To this end, the considered one-day time horizon is discretized into $T$ time intervals (referred to as time-slots) indexed by $\timelist$, each of duration $\dt$. The introduced framework is very general and considers several sources of uncertainty like the first $a_i$ and last $d_i$ time-slots EV $i$ is connected, the $i$-th battery energy content at EV arrival $e_i^0$, and the TSO service signal $\omega_k \in [-1,1]$ modeling if and how much of the offered upward/downward services will be requested by the TSO. These are all random quantities as they are not known during day-ahead planning.

Next, we briefly review the EV modeling and cost terms introduced in~\cite{Vignali2022}.

\subsubsection{Electric Vehicles Modeling} \label{sec:EV_model}
In~\cite{Vignali2022}, each EV is modeled as a battery
\begin{equation} \label{eq:bat_dyn}
	e_{k+1,i} = \alpha_i \, e_{k,i} + \dt \, \eta_{k,i} \, p_{k,i}, \qquad k \in [a_i,d_i],
\end{equation}
where $e_{k,i}$ is the energy content at the beginning of time-slot $k$, $p_{k,i}$ denotes the average charging ($p_{k,i} > 0$) or discharging ($p_{k,i} < 0$) power during the time-slot $k$, $\alpha_i \in (0,1]$ models self-discharging losses, and 
	\begin{equation} \label{eq:bat_eff}
		\eta_{k,i} =
		\begin{cases}
			\eta_i^+						& p_{k,i} \ge 0 \\
			\textstyle\frac{1}{\eta_i^-}	& p_{k,i} < 0
		\end{cases}
	\end{equation}
models charging/discharging losses, $\eta_i^+,\eta_i^- \in (0,1]$ being the charging/discharging efficiencies.

The battery energy content $e_{k,i}$ always stays within a minimum $e_i^{\min} > 0$ and a maximum $e_i^{\max} > 0$ value and therefore
\begin{equation} \label{eq:lim_soc}
	e_i^{\min} \le e_{k,i} \le e_i^{\max}
\end{equation}
must hold for any time-slot $k \in [a_i,d_i]$ in which EV $i$ is connected to the charging station.

Similarly, individual and aggregate power exchanges are constrained as
\begin{subequations} \label{eq:lim_rate}
	\begin{align}
		&p_{k,i} \in [-p_i^{\max}, p_i^{\max}]		&k \in [a_i,d_i] \label{eq:lim_rate_bound} \\
		&p_{k,i} = 0								&k \not\in [a_i,d_i] \label{eq:lim_rate_zero}
	\end{align}
\end{subequations}
since each EV has a maximum power exchange $p_i^{\max}$ when connected and its power exchange must be zero when disconnected, and 
\begin{equation} \label{eq:lim_rate_all}
	-p^{\max} \le \vehsum p_{k,i} \le p^{\max},
\end{equation}
for all time-slots, as the charging stations are all connected to the same point of exchange with the grid, which can withstand a maximum power exchange equal to $p^{\max}$.

In most cases, a minimum battery energy content at departure is required by the user. This can be easily taken into account by enforcing the constraint
\begin{equation} \label{eq:ref_soc}
	e_{d_i+1,i} \ge e_i^\circ,
\end{equation}
where $e_i^\circ \in [e_i^{\min},e_i^{\max}]$ is EV $i$ desired energy at departure.

\subsubsection{Cost Terms} \label{sec:cost_terms}
In order to charge the EVs, the aggregator has to buy energy on the market and it can do so both on the Day-Ahead Market (DAM) or on the Ancillary Services Market (ASM). Let $p_{k,i} = p_{k,i}^\dam + p_{k,i}^\asm$, $p_{k,i}^\dam$ and $p_{k,i}^\asm$ being the portion of power bought in DAM and in ASM, respectively.

At any time-slot $k$, buying an energy unit on the DAM costs $c_k^{e+}$ to the aggregator, while selling energy to the grid pays $c_k^{e-} < c_k^{e+}$ per energy unit. The aggregator buys energy whenever the net power requested by all EVs is positive, and sells energy otherwise. The cost incurred for the DAM over the entire horizon is thus given by
\begin{equation} \label{eq:cost_dam}
	c^\dam = \timesum c_k^{e+} \posp{\vehsum \dt p_{k,i}^\dam} - c_k^{e-} \negp{\vehsum \dt p_{k,i}^\dam},
\end{equation}
where $\posp{v} = \max\{v,0\}$ denotes the positive part of its argument and $\negp{v} = \posp{-v}$ its negative part.

As for the ancillary services, since they are typically divided into upward and downward services, it is convenient to express $p_{k,i}^\asm$ as
\begin{equation} \label{eq:pki_asm}
	p_{k,i}^\asm = \posp{\omega_k} s_{k,i}^+ - \negp{\omega_k} s_{k,i}^-,
\end{equation}
where $s_{k,i}^+ \ge 0$ and $s_{k,i}^- \ge 0$ are the maximum power variations offered by EV $i$ in time-slot $k$ for the downward and upward services respectively, and $\omega_k$ is the uncertain service signal sent by the TSO. 

An energy unit bought (as a downward service) on the ASM costs $c_k^{s+} < c_k^{e+}$, while an energy unit sold (as an upward service) pays $c_k^{s-} > c_k^{e+}$, leading to the following total cost incurred by the aggregator
\begin{equation} \label{eq:cost_asm}
	c^\asm = \timesum c_k^{s+} \vehsum \dt \posp{\omega_k} s_{k,i}^+ - c_k^{s-} \vehsum \dt \negp{\omega_k} s_{k,i}^-,
\end{equation}
where the sign and magnitude of $\omega_k \in [-1,1]$ determine whether an upward ($\omega_k < 0$) or downward ($\omega_k > 0$) will be requested by the TSO and by which extent, or if no service will be requested ($\omega_k = 0$), for each time-slot $k$.

Depending on the situation, the aggregator may want to charge/pay the EV owners for recharging/discharging their vehicles. In such cases, the aggregator will receive $c_k^{v+} > c_k^{e+}$ per energy unit used for charging an EV and will pay $c_k^{v-} > c_k^{v+}$ to the EV owner for each energy unit discharged\footnote{Costs for battery degradation can be accounted for in $c_k^{v-}$,~\cite{hoke2014accounting}}, thus having the following additional cost term
\begin{equation} \label{eq:cost_veh}
	c^\veh = \timesum \vehsum  c_k^{v-} \negp{\dt p_{k,i}} - c_k^{v+} \posp{\dt p_{k,i}}.
\end{equation}

\subsubsection{Optimal Planning} \label{sec:formulation}
Unfortunately, an optimization problem involving the introduced constraints and cost terms would be ill-posed due to their dependency from the uncertain parameters $a_i$, $d_i$, $e_i^0$, and $\omega_k$, collectively referred to as $\delta$. \cite{Vignali2022} proposes to adopt a robust paradigm to enforce the constraints and an expectation paradigm for the cost function. Accordingly, we will focus on the following problem
\begin{align*}
	\min_{p_{k,i}^\dam, s_{k,i}^+, s_{k,i}^-} \quad
		&c^\dam + \Exp[c^\asm + c^\veh] \hspace*{-10em}					&& \tag{$\cP$} \label{eq:opt_det} \\
	\text{subject to:} \quad
		&\eqref{eq:lim_rate_all} 										&& \forall k,~\forall \delta \\
		&\eqref{eq:lim_soc}, \eqref{eq:lim_rate}, \eqref{eq:ref_soc}	&& \forall i,~\forall \delta \\
		&s_{k,i}^+, s_{k,i}^- \ge 0										&& \forall i,~\forall k.
\end{align*}
More specifically, we will consider two business models: 1) \emph{Free-Charge (FC)}, without the $c^\veh$ term, representative of a company willing to provide the recharge service to its employees and, ii) \emph{Paid Charge (PC)}, with the $c^\veh$ term, in case EVs charging is the core business of the parking lot owner.

\section{Profitability Analysis} \label{sec:analysis}
We are interested in providing conditions under which the provision of ancillary services is profitable for an aggregator of EVs adopting either the FC or PC business model introduced above. Numerical investigations, despite being informative, can hardly give an overall picture on profitability, as results are masked by the complexity of problem~\ref{eq:opt_det} and by the uncertainty affecting the simulation of optimal control policies. Therefore, in this section, we first simplify the framework in order to, then, derive precise necessary and sufficient conditions for profitability.

\subsection{Framework Reduction} \label{sec:simplifying_ass}
We impose the following simplifying assumptions. We consider the optimization of one vehicle ($N = 1$ and we drop the subscript $i$), as multiple vehicles can be considered, to some extent, a unique ``big'' vehicle. We set $\eta_k = \alpha = 1$ as they are typically close to unity. All costs are time-invariant (we drop the subscript $k$). Since the costs are time-invariant, we can consider a unique time-slot ($T = 1$) lasting $\dt = 24$h and reduce the analysis to energy considerations. Since $T = 1$, we set $a_i = a = d_i = d = 0$, which are now deterministic. We consider only the TSO request $\omega_k = \omega$ as uncertain quantity since the initial energy $e_i^0 = e_0 \in [e^{\min},e^\circ]$ is only affecting the constraints.

Under these assumptions, we have
\begin{align*}
	p &= p^\dam +  s^+ \posp{\omega} - s^- \negp{\omega} \\
	e_1 &= e_0 + \dt p \\
		&= e_0 + \dt p^\dam +  \dt s^+ \posp{\omega} - \dt s^- \negp{\omega}
\end{align*}
and
\begin{equation} \label{eq:red_costs}
\begin{aligned}
	&c^\dam = c^{e+} \posp{\dt p^\dam} - c^{e-} \negp{\dt p^\dam} \\
	&c^\asm = c^{s+} \posp{\omega} \dt s^+ - c^{s-} \negp{\omega} \dt s^- \\
	&c^\veh = c^{v-} \negp{\dt p} - c_k^{v+} \posp{\dt p},
\end{aligned}
\end{equation}
and~\ref{eq:opt_det} becomes
\begin{align*}
	\min_{p^\dam, s^+, s^-} \quad
		&c^\dam + \Exp[c^\asm + c^\veh]								\tag{$\cP_s$} \label{eq:opt_det_simplified} \\
	\text{subject to:} \quad
		&e^{\min} \le e^\circ \le e_1 = e_0 + \dt p \le e^{\max}	&& \forall \omega \\
		&-p^{\max} \le p \le p^{\max}								&& \forall \omega \\
		&s^+, s^- \ge 0.
\end{align*}
Since we typically have $\dt p^{\max} \gg e^{\max}$ (over the entire horizon we can charge the EV fully), the power constraints are redundant and we are left with only energy quantities in~\ref{eq:opt_det_simplified}. Let $\edam = \dt p^\dam$, $\ep = \dt s^+$, and $\en = \dt s^-$, the robust counterpart of $e^\circ \le e_0 + \dt p \le e^{\max}$ for all $\omega$ affecting $\dt p$ is given by $$\Dr + \en \le \edam \le \Dmax - \ep,$$ where $\Dr = e^\circ - e_0$ and $\Dmax = e^{\max} - e_0$, and it implies $\edam \ge 0$. Problem~\ref{eq:opt_det_simplified} can thus be further reduced to
\begin{align*}
	\min_{\edam, \ep, \en \ge 0} \quad
		&c^\dam + \Exp[c^\asm + c^\veh]			\tag{$\cP^\asm$} \label{eq:opt_FC_PC_asm} \\
	\text{subject to:} \quad
		&\Dr + \en \le \edam \le \Dmax - \ep.
\end{align*}

Before analyzing~\ref{eq:opt_FC_PC_asm} it is worth recalling some inequalities involving the unitary energy prices
\begin{subequations} \label{eq:cost_ineqs}
\begin{align}
	c^{v-} > c^{v+} &> c^{e+} > c^{e-}, \label{eq:cost_ineqs_dam_veh} \\
	c^{s-} &> c^{e+} > c^{s+}. \label{eq:cost_ineqs_asm}
\end{align}
\end{subequations}

\subsection{Profitability Conditions: Free Charge} \label{sec:profitability_FC}
Let us consider the FC case first, where the cost term $c^{\veh}$ is absent. To assess whether offering ancillary services is profitable or not, consider the optimal solution when such services are not offered. This entails solving the following optimization problem
\begin{align*}
	\min_{\edam \ge 0} \quad
		&c^\dam						\tag{$\cP_{FC}^0$} \label{eq:opt_FC_baseline} \\
	\text{subject to:} \quad
		&\Dr \le \edam \le \Dmax,
\end{align*}
whose optimal solution is $\edam = \Dr$ since $c^{\dam} = c^{e+} \edam$ due to $\edam \ge 0$ and $c^{e+} > 0$.

If we now introduce the ancillary service provision, we are back to~\ref{eq:opt_FC_PC_asm} without $c^{\veh}$. Clearly, $\edam = \Dr$ with $\ep = 0$ and $\en = 0$ is feasible for~\ref{eq:opt_FC_PC_asm} and yields the same cost
\begin{equation*}
	J_{FC}^0 = c^{e+} \posp{\edam} - c^{e-} \negp{\edam} = c^{e+} \edam = c^{e+} \Dr,
\end{equation*}
the second and third equality being due to $\edam = \Dr > 0$.

Therefore, for~\ref{eq:opt_FC_PC_asm} without $c^{\veh}$ to have a different solution there must exist a triplet $\et = \Dr + v$, $\ep \ge 0$, and $\en \ge 0$ satisfying the constraint of~\ref{eq:opt_FC_PC_asm}, i.e.,
\begin{equation} \label{eq:feasibility_FC_asm}
	\Dr + \en \le \Dr + v \le \Dmax - \ep
\end{equation}
and achieving a better cost. Constraint~\eqref{eq:feasibility_FC_asm} together with non-negativity of $\ep$ and $\en$ implies the following chain of inequalities
\begin{equation} \label{eq:feasibility_FC_asm_chain}
	0 \le \en \le v \le \Dmax - \Dr - \ep \le \Dmax - \Dr,
\end{equation}
and the cost associated to the new solution is
\begin{align}
	J_{FC}^\asm
		&= c^{e+} \posp{\et} - c^{e-} \negp{\et} + c^{s+} \Exp^+ \ep - c^{s-} \Exp^- \en \nonumber \\
		&= c^{e+} \Dr + c^{e+} v + c^{s+} \Exp^+ \ep - c^{s-} \Exp^- \en \nonumber \\
		&= J_{FC}^0 + \underbrace{c^{e+} v + c^{s+} \Exp^+ \ep - c^{s-} \Exp^- \en}_{J_{FC}^\Delta} \label{eq:new_cost_FC}
\end{align}
where $\Exp^+ = \Exp[\posp{\omega}]$ and $\Exp^- = \Exp[\negp{\omega}]$, the second equality is due to $\et = \Dr + v \ge 0$ since $v \ge 0$ by~\eqref{eq:feasibility_FC_asm_chain}, and the last equality is by definition of $J_{FC}^0$. We thus need to analyze the sign of $J_{FC}^\Delta$.

For any triplet $(v,\ep,\en)$ satisfying~\eqref{eq:feasibility_FC_asm_chain}, we have
\begin{align*}
	J_{FC}^\Delta
		&= c^{e+} v + c^{s+} \Exp^+ \ep - c^{s-} \Exp^- \en \\
		&\ge c^{e+} v - c^{s-} \Exp^- \en \\
		&\ge (c^{e+} - c^{s-} \Exp^-) \en \\
		&\ge \min\{ 0, (c^{e+} - c^{s-} \Exp^-) (\Dmax - \Dr) \},
\end{align*}
where the first inequality holds for any $\ep \ge 0$ (with $\ep = 0$ as edge-case), the second inequality holds for any $v \ge \en$ (with $v = \en$ as edge-case), and the last inequality holds for any $\en$ such that $0 \le \en \le \Dmax - \Dr$, with $\en = 0$ or $\en = \Dmax - \Dr$ as edge-cases, each one yielding the respective term inside the minimum. Since $\Dmax - \Dr > 0$, then, recalling $\et = \Dr + v$, we have the following edge-cases:
\begin{align*}
	&
	\begin{cases}
		v = 0 \\
		\et = \Dr \\
		\ep = 0 \\
		\en = 0 \\
		J_{FC}^\Delta = 0
	\end{cases}
	&&
	\iff
	&&
	c^{e+} > c^{s-} \Exp^-
	\\
	&
	\begin{cases}
		v = \Dmax-\Dr \\
		\et = \Dmax \\
		\ep = 0 \\
		\en = \Dmax-\Dr \\
		J_{FC}^\Delta < 0
	\end{cases}
	&&
	\iff
	&&
	c^{e+} < c^{s-} \Exp^-
\end{align*}
Therefore, if $c^{e+} > c^{s-} \Exp^-$, then $J_{FC}^\Delta \ge 0$ for any feasible alternative solution, hence $\edam = \Dr$ remains the optimal solution. Otherwise, if $c^{e+} < c^{s-} \Exp^-$, then choosing $\et = \Dr + v = \Dmax$, $\ep = 0$, and $\en = \Dmax - \Dr$ yields $J_{FC}^\Delta = (c^{e+} - c^{s-} \Exp^-) (\Dmax - \Dr) < 0$ and offering (upward) services is profitable. Note that since $(c^{e+} - c^{s-} \Exp^-) (\Dmax - \Dr) < J_{FC}^\Delta$ for any feasible solution, we have that $\et = \Dmax$, $\ep = 0$, and $\en = \Dmax - \Dr$ is actually the optimal solution of~\ref{eq:opt_FC_PC_asm} without $c^{\veh}$, so offering downward services is never convenient and the obtained condition is both necessary and sufficient for profitability.

The condition is also intuitive as providing upward services is convenient only if their expected revenue $c^{s-} \Exp^-$ per unit is greater than the cost $c^{e+}$ of buying an energy unit in the DAM. Note also how the optimal strategy is to offer as an upward service only the quantity $\Dmax - \Dr = e^{\max} - e^\circ$ as offering more energy does not guarantee to satisfy the final energy constraint $e_1 \ge e^\circ$.

\subsection{Profitability Conditions: Paid Charge} \label{sec:profitability_PC}
Let us now focus on the PC case. As before, consider first the optimal solution when services are not offered. This entails solving the problem
\begin{align*}
	\min_{\edam \ge 0} \quad
		&c^\dam + c^\veh			\tag{$\cP_{PC}^0$} \label{eq:opt_PC_baseline} \\
	\text{subject to:} \quad
		&\Dr \le \edam \le \Dmax,
\end{align*}
where $c^\veh$ is now a deterministic cost since $\dt p = \edam > 0$ when $\ep = \en = 0$. Moreover, $c^\dam + c^\veh = (c^{e+}-c^{v+}) \edam$ and, since $c^{e+}-c^{v+} < 0$ by~\eqref{eq:cost_ineqs_dam_veh}, then the optimal solution of~\ref{eq:opt_PC_baseline} is $\edam = \Dmax$.

If we now introduce the ancillary service provision, we are back to~\ref{eq:opt_FC_PC_asm}. Clearly, $\edam = \Dmax$ with $\ep = 0$ and $\en = 0$ is feasible for~\ref{eq:opt_FC_PC_asm} and yields the same cost
\begin{align*}
	J_{PC}^0
		&= (c^{e+} - c^{v+}) \posp{\edam} - (c^{e-} - c^{v-}) \negp{\edam} \\
		&= (c^{e+} - c^{v+}) \edam = (c^{e+} - c^{v+}) \Dmax,
\end{align*}
equalities being due to $\edam = \Dmax > 0$.

Therefore, for~\ref{eq:opt_FC_PC_asm} to have a different solution there must exist a triplet $\et = \Dmax + v$, $\ep \ge 0$, and $\en \ge 0$ satisfying the constraint of~\ref{eq:opt_FC_PC_asm}, i.e.,
\begin{equation} \label{eq:feasibility_PC_asm}
	\Dr + \en \le \Dmax + v \le \Dmax - \ep
\end{equation}
and achieving a better cost. Constraint~\eqref{eq:feasibility_PC_asm} together with non-negativity of $\ep$ and $\en$ implies the following chain of inequalities
\begin{equation} \label{eq:feasibility_PC_asm_chain}
	- (\Dmax - \Dr) \le \en - (\Dmax - \Dr) \le v \le - \ep \le 0,
\end{equation}
and the cost associated to the new solution is
\begin{align}
	J_{PC}^\asm
		&= c^{e+} \posp{\et} - c^{e-} \negp{\et} + c^{s+} \Exp^+ \ep - c^{s-} \Exp^- \en \nonumber \\
			&~~~~ + c^{v-} \Exp \left[ \negp{\et + \ep \posp{\omega} - \en \negp{\omega}} \right] \nonumber \\
			&~~~~ - c^{v+} \Exp \left[ \posp{\et + \ep \posp{\omega} - \en \negp{\omega}} \right] \nonumber \\
		&= c^{e+} \Dmax + c^{e+} v + c^{s+} \Exp^+ \ep - c^{s-} \Exp^- \en \nonumber \\
			&~~~~ - c^{v+} \Exp[\Dmax + v + e^+ \posp{\omega} - e^- \negp{\omega}] \nonumber \\
		&= (c^{e+} - c^{v+}) \Dmax + (c^{e+} - c^{v+}) v \nonumber \\
			&~~~~ + (c^{s+} - c^{v+}) \Exp^+ \ep + (c^{v+} - c^{s-}) \Exp^- \en \nonumber \\
		&= J_{PC}^0 + J_{PC}^\Delta \label{eq:new_cost_PC}
\end{align}
where the second equality is due to $\et = \Dmax + v \ge 0$ since $v \ge -(\Dmax-\Dr)$ by~\eqref{eq:feasibility_PC_asm_chain} together with $\et + \ep \posp{\omega} - \en \negp{\omega} \ge \Dmax + v - \en \ge \Dr$ by~\eqref{eq:feasibility_PC_asm_chain}, non-negativity of $\ep$ and $\en$, and the fact that $\omega \in [-1,1]$. The third equality is due to linearity of the expected value operator and the last equality uses the definition of $J_{PC}^0$ and $J_{PC}^\Delta = (c^{e+} - c^{v+}) v + (c^{s+} - c^{v+}) \Exp^+ \ep + (c^{v+} - s^-) \Exp^- \en$. Similarly to the FC case, we need to analyze the sign of $J_{PC}^\Delta$.

To ease the notation, let $g^0 = c^{v+} - c^{e+} > 0$, $g^+ = \Exp^+ (c^{v+} - c^{s+}) > 0$, and $g^- = \Exp^- (c^{s-} - c^{v+})$, inequalities being due to~\eqref{eq:cost_ineqs}. For any triplet $(v,\ep,\en)$ satisfying~\eqref{eq:feasibility_PC_asm_chain}, we have
\begin{align}
	J_{PC}^\Delta
		&= - g^0 v - g^+ \ep - g^- \en \nonumber \\
		&\ge (g^+ - g^0) v - g^- \en \nonumber \\
		&\ge \min \{ 0, (g^0 - g^+) (\Dmax - \Dr), - g^- (\Dmax - \Dr) \},
\end{align}
where the first inequality is due to $-\ep \ge v$ (with $\ep = -v$ as edge-case) and the second inequality is given by the fact that, due to $\en \le v + \Dmax - \Dr$ with $v \le 0$ and $\en \ge 0$, we are left with three possible edge-cases: $v = \en = 0$ or $v = -(\Dmax-\Dr)$ and $\en = 0$ or $v = 0$ and $\en = \Dmax-\Dr$ (with $\ep = -v$ in all cases), each one yielding the respective term inside the minimum. Since $\Dmax-\Dr > 0$, then, recalling $\et = \Dmax + v$, we have the following edge-cases:
\begin{align*}
	&
	\begin{cases}
		v = 0 \\
		\et = \Dmax \\
		\ep = 0 \\
		\en = 0 \\
		J_{PC}^\Delta = 0
	\end{cases}
	&&
	\iff
	&&
	\begin{cases}
		g^+ - g^0 < 0 \\
		g^- < 0
	\end{cases}
	\\
	&
	\begin{cases}
		v = -(\Dmax-\Dr) \\
		\et = \Dr \\
		\ep = \Dmax-\Dr \\
		\en = 0 \\
		J_{PC}^\Delta < 0
	\end{cases}
	&&
	\iff
	&&
	\begin{cases}
		g^+ - g^0 > 0 \\
		g^- < g^+ - g^0
	\end{cases}
	\\
	&
	\begin{cases}
		v = 0 \\
		\et = \Dmax \\
		\ep = 0 \\
		\en = \Dmax-\Dr \\
		J_{PC}^\Delta < 0
	\end{cases}
	&&
	\iff
	&&
	\begin{cases}
		g^- > 0 \\
		g^- > g^+ - g^0
	\end{cases}
\end{align*}
Similarly to the FC case, since each edge-case achieves the minimum of $J_{PC}^\Delta$ under the respective conditions on $g^0$, $g^+$, and $g^-$, we have that each edge-case is actually the optimal solution of~\ref{eq:opt_FC_PC_asm} under the corresponding conditions. Therefore the above conditions on $g^0$, $g^+$, and $g^-$ are both necessary and sufficient for profitability of upward/downward services.

\begin{figure}[t]
	\centering
	\includegraphics[width=0.8\columnwidth]{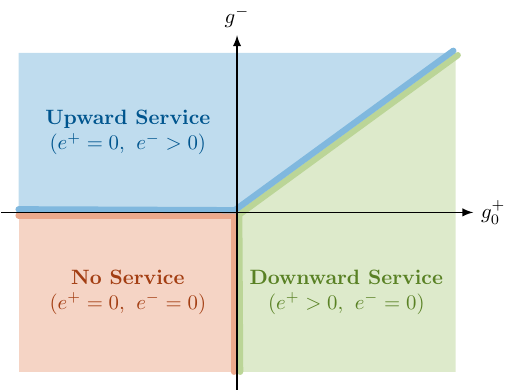}
	\caption{Partition of the $(g_0^+,g^-)$ plane induced by the profitability conditions in the PC case. Each region represent under which conditions the service is profitable (hence offered).}
	\label{fig:PC_profit_regions}
\end{figure}

Now let us notice that $g^0 = c^{v+} - c^{e+}$ represents the marginal gain of buying an energy unit in the DAM and selling it to the vehicle, $g^+ = \Exp^+ (c^{v+} - c^{s+})$ represents the marginal gain of buying an energy unit in the ASM (downward service) and selling it to the vehicle, $g^- = \Exp^- (c^{s-} - c^{v+})$ represents the marginal gain of selling an energy unit in the ASM (upward service) instead of selling it to the vehicle (i.e., the actual (expected) gain in offering the upward service), and $g_0^+ = g^+ - g^0$ represents the marginal gain of buying an energy unit in the ASM (downward service) instead of in the DAM to sell to the vehicle (i.e., the actual (expected) gain in offering the downward service). Given the preceding observations, we need to focus on $g_0^+$ and $g^-$ only, and the above optimality conditions becomes also intuitive.

To aid the interpretation, we report in Figure~\ref{fig:PC_profit_regions} the partition of the $(g_0^+,g^-)$ plane induced by the edge-cases conditions. In the III quadrant $g_0^+ < 0$ and $g^- < 0$, meaning that there is no advantage in offering a downward service w.r.t buying energy on the DAM and there is no gain in offering an upward service instead of selling energy to the vehicle, hence the best strategy is to fully charge the vehicle buying from the DAM and not to offer any service. If $g_0^+ > 0$ and $g^- < g_0^+$, then offering a downward service instead of buying energy on the DAM is profitable, and offering an upward service is either not profitable (IV quadrant) or not as profitable as a downward one (lower part of I quadrant), hence the best strategy is to fully charge the vehicle buying $\Dr$ from DAM and $\Dmax-\Dr$ from ASM as a downward service. Finally, if $g^- > 0$ and $g^- > g_0^+$, then offering an upward service w.r.t. selling energy to the vehicle is profitable and offering a downward service is either not profitable (II quadrant) or not as profitable as an upward one (upper part of I quadrant), hence the optimal solution is to fully charge the vehicle buying from the DAM and offer $\Dmax-\Dr$ as an upward service.

Finally, note that the cost coefficients involved in the analysis are $c^{v+}$, $c^{e+}$, $c^{s+}$, and $c^{s-}$, while $c^{e-}$ and $c^{v-}$ do not appear. This is due to the fact that with a single time-slot, the vehicle cannot be discharged, hence $p^\dam$ and $p$ are always positive. We expect these cost coefficients to pop up in the multiple time-slot case, whose analysis is left as a future research effort.

\subsection{Successive markets and unbalance} \label{sec:unbalancing}
By solving~\ref{eq:opt_det}, the aggregator computes the optimal amount of energy to buy or sell on the energy and ancillary services markets. According to the setting in~\cite{Vignali2022}, this decision is taken at day $t-1$ (i.e., the day ahead) and implemented as-is in day $t$ (i.e., the day after). However, in practice, as time goes by in day $t$, the aggregator can update its profile for the remaining part of day $t$ by buying or selling energy on the so-called infra-day markets, or can even choose not to follow the scheduled profile, thus unbalancing the grid. This possibility is currently not exploited in~\cite{Vignali2022}, but it could be included by making $p_{k,i}$ dependent on the TSO service signal $\omega_s$, $s = 0,\dots,k-1$, with an additive ``disturbance feedback'' term $K_{k,s,i} \omega_s$. The gain $K_{k,s,i}$ will still be optimized the day ahead, but it will produce a power profile $p_{k,i}$ which, at day $t$, changes according to the actual realization of the TSO service signal up to time-slot $k-1$, which will be known at time-slot $k$ of day $t$. This modification would increase the profitability of the ancillary services, as the following example clarifies.

Consider offering an upward ancillary service in the PC case: the analysis in Section~\ref{sec:profitability_PC} shows that it is profitable if and only if the return of selling to the ancillary services market is greater than selling to the user. When this is the case, the optimal strategy is to sell the entire (allowed by the user) vehicle capacity to the market. However, if the aggregator was allowed to unbalance, he could sell the entire capacity of the vehicles as an upward service, wait for the request by the TSO, and - afterwards - absorb the same amount of energy that has been requested (i.e. a disturbance feedback with unitary gain), eventually incurring in unbalance costs. As long as the unbalance cost is sufficiently small, this strategy would lead to a greater profitability, since the vehicles would depart fully charged independently on the actual realization of the TSO service signal.

\section{Case Study} \label{sec:example}
Here we make some simulations using the full-fledged framework in~\cite{Vignali2022} to show that the conditions found in Section~\ref{sec:analysis} are sufficient also for the general case, without simplifying assumptions. Due space limits, we investigate the paid charge case only.

\begin{figure}
	\centering
	\includegraphics[width=.8\columnwidth]{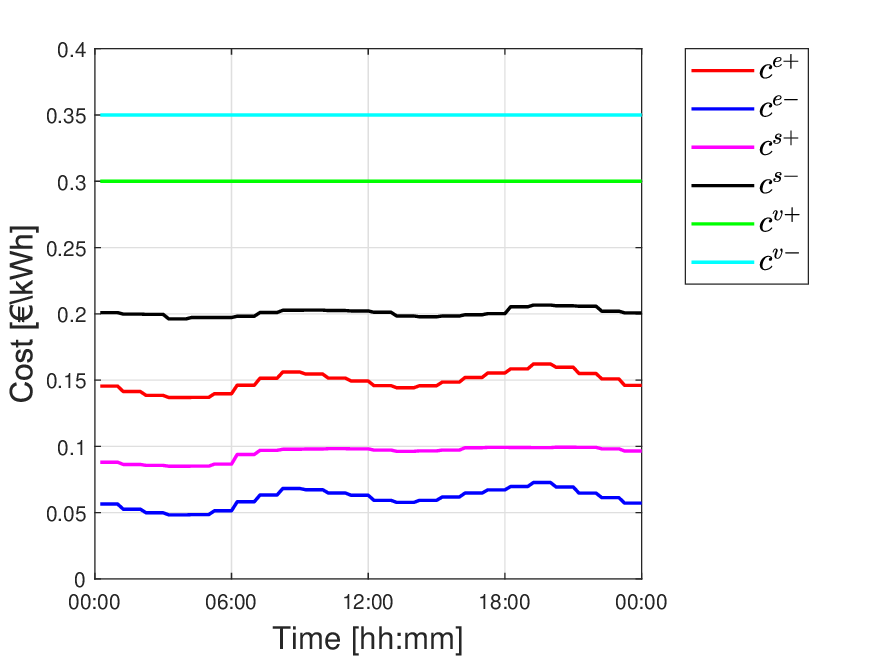}
	\caption{Day-ahead market, ancillary service market, and vehicle charging/discharging prices. \label{fig:prices}}
\end{figure}

We consider the case of a company parking lot composed of $N = 100$ slots, each assigned to a single user indexed with $i$. The 24 hours time horizon is discretized into $T = 96$ time slots of $\dt = 15$ minutes each. Vehicle $i$ arrives uniformly at random between 6:00 AM and 8:00 AM	and leaves uniformly at random between 4:00 PM and 8:00 PM. For each vehicle $i$, we set $\eta^{+}_i = \eta^{-}_i = 0.97$, $p^{\max}_i = -p^{\min}_i = 22$ kW, $e_i^{\min} = 0$ kWh, and $e_i^{\circ} = 0.7e_i^{\max}$, with $e_i^{\max} \in [40, \ 70]$ kWh and $e_i^0 \in [0.1,\ 0.3]e_i^{\max}$ kWh extracted at random according to a uniform distribution. The maximum power that can be exchanged with the grid is set to $p^{\max} = -p^{\min} = 600$ kW. The energy unitary prices\footnote{Real Italian market data (see~\cite{GME}) in 2018.} are shown in Figure~\ref{fig:prices} while the acceptance probabilities for the ASM are set to $\pi^{+}_k = 0.6$ and $\pi^{-}_k = 0.1$, for downward and upward services respectively.

\begin{figure*}
	\centering
	\includegraphics[width=\textwidth]{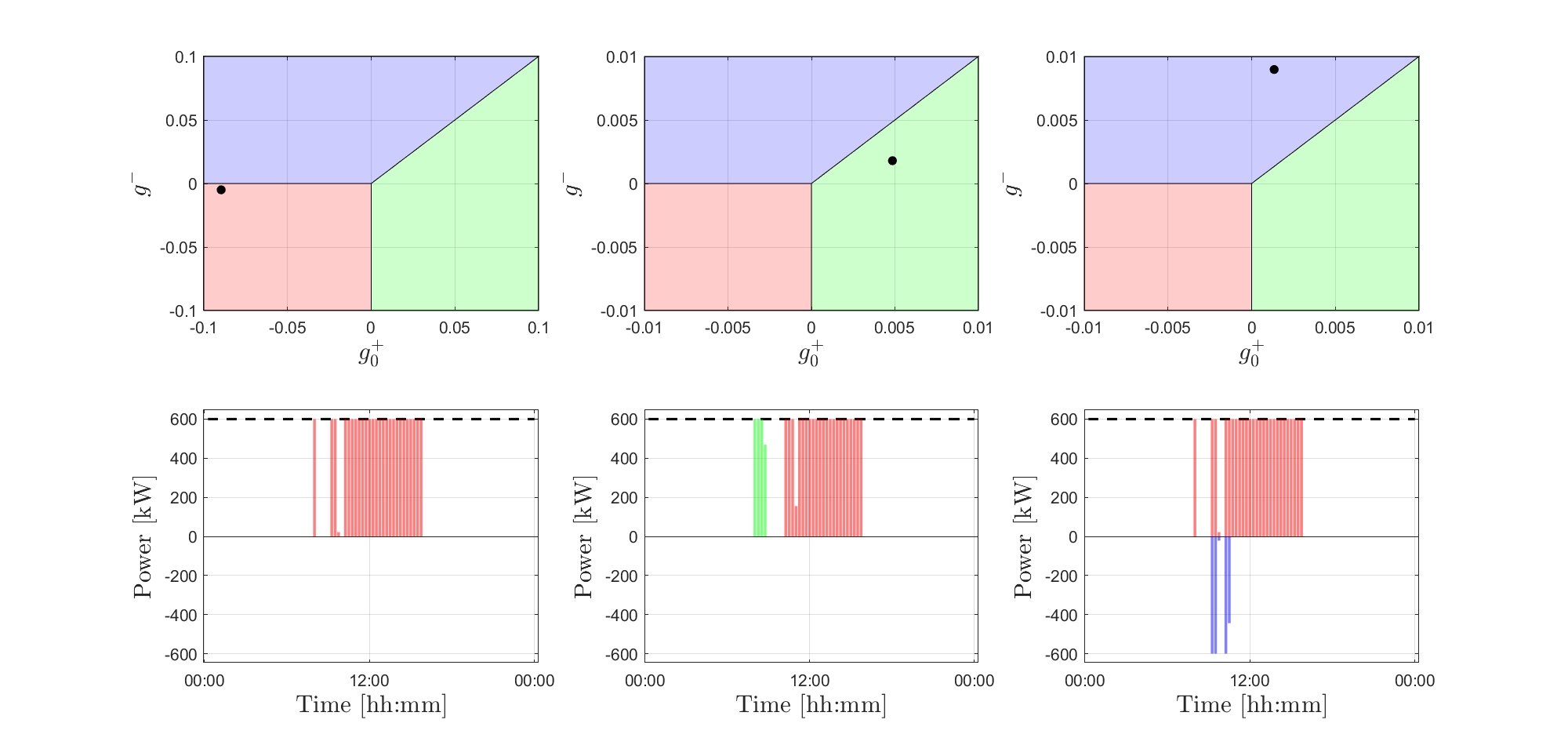}
	\caption{Top: ($g_0^+$,$g^-$) plane. Black dots are obtained by considering day-averaged prices when computing $g_0^+$ and $g^-$. Bottom: Simulation results for the vehicle fleet -- power exchange profile with the main grid. Different colors denote energy bought on different markets: day-ahead market (i.e., no service) (red), downward service (green), upward service (blue). Dashed line denotes the aggregate power limit. From left to right: no service case, downward service case, upward service case.}
	\label{fig:points}
\end{figure*}

The results of our numerical investigation are summarized in Figure~\ref{fig:points} where we report the optimal charging power profile $\vehsum p_{k,i}$, $k = 0,\dots,T-1$, of the aggregator (bottom plots) and the corresponding working point in the ($g_0^+$,$g^-$) plane (top plots), using day-averaged prices to compute $g_0^+$ and $g^-$, in three different cases (left to right). With the parameter values introduced before, the optimal charging policy consists in buying all the energy on the DAM and use it to fully charge the vehicles, see Figure~\ref{fig:points} (bottom left) where the power bought by the aggregator is red (i.e., bought from DAM) at all time slots. Indeed, by considering the average energy prices and computing $g_0^+$ and $g^-$, we fall into the no-service case, as shown in Figure~\ref{fig:points} (top left). If we now reduce the vehicle prices and set them to $c^{v+} = 0.165$ \euro/kWh and $c^{v-} = 0.18$ \euro/kWh, then we can see that downward ancillary services become profitable, as shown in Figure~\ref{fig:points} (top center), where the power bought by the aggregator is partially in green (i.e., bought from ASM), and in Figure~\ref{fig:points} (bottom center), where the average energy prices map into a point in the downward service area. Note that in this case also upward services are profitable ($g^- > 0$) but not as profitable as the downward ones. Finally, we raised the acceptance probability of upward services to $\pi^{-} = 0.5$ (and reduced $\pi^{+}$ to $0.5$) to make upward services more profitable and we obtained the optimal aggregator power profile in Figure~\ref{fig:points} (bottom right), where it can be seen that some time slots are blue (i.e., power bought from ASM). Accordingly, the average energy prices maps into a point falling into the upward service case, see Figure~\ref{fig:points} (top right).

These three cases show that our results, despite being derived based on strong simplifying assumptions, are actually valid also when we consider the more complicated framework along with different sources of uncertainty.

\section{Conclusion} \label{sec:conclusion}

A theoretical profitability analysis with reference to a recently proposed framework for V2G optimal operation in presence of uncertainty has been proposed. Under simplifying assumptions, some necessary and sufficient conditions for profitability have been derived and then numerically proved correct also for the general case, via simulation, by considering a company parking lot as a realistic application scenario. Backing the general case with rigorous theoretical findings is left as a future research effort.

The performed numerical investigation also showed how this profitability analysis can be used to study how the variation of some parameters, such as the service acceptance probability or even some costs, impact on the provision of ancillary services. Finally, based on the obtained conditions, we also provided hints on how to make ancillary services more profitable through the participation to other existing markets, such as the so-called infra-day markets, or even by choosing to unbalance the grid. This aspect will be the topic of further research activities.

\small
\bibliography{biblio}

%

\end{document}